\documentclass{article}
\usepackage{ijcai22}

\usepackage{times}
\usepackage{soul}
\usepackage{url}
\usepackage[hidelinks]{hyperref}
\usepackage[utf8]{inputenc}
\usepackage[small]{caption}
\usepackage{graphicx}
\usepackage{amsmath}
\usepackage{amsthm}
\usepackage{booktabs}
\usepackage{algorithm}
\usepackage{algorithmic}
\usepackage{multirow}
\title{Efficient Pressure: Improving efficiency for signalized intersections}

\author{
Qiang Wu$^1$ \and 
Liang Zhang$^2$\footnote{Liang Zhang is the corresponding author.}\and 
Jun Shen$^3$\and
Linyuan Lü$^1$\and 
Bo Du$^4$\and 
Jianqing Wu$^3$\\
\affiliations
$^1$Institute of Fundamental and
    Frontier Sciences, University of Electronic Science and Technology of
    China, Chengdu 611731, China\\
$^2$ School of Life Sciences, Lanzhou University, Lanzhou 730000, China\\
$^3$School of Computing and Information Technology, University of Wollongong, Wollongong, Australia\\
$^4$SMART Infrastructure Facility, University of Wollongong, Wollongong, Australia\\
\emails
qiang502.wu@gmail.com,
zhangliang18@lzu.edu.cn,
jshen@uow.edu.au,
linyuan.lv@uestc.edu.cn,
bdu@uow.edu.au,
jw937@uowmail.edu.au
}

\begin{document}

\maketitle

\begin{abstract}
 Since conventional approaches could not adapt to dynamic traffic conditions, reinforcement learning (RL) has attracted more attention to help solve the traffic signal control (TSC) problem. However, existing RL-based methods are rarely deployed considering that they are neither cost-effective in terms of computing resources nor more robust than traditional approaches, which raises a critical research question: how to construct an adaptive controller for TSC with less training and reduced complexity based on RL-based approach?
To address this question, in this paper, we (1) innovatively specify the traffic movement representation as a simple but efficient pressure of vehicle queues in a traffic network, namely efficient pressure (EP); (2) build a traffic signal settings protocol, including phase duration, signal phase number and EP for TSC; (3) design a TSC approach based on the traditional max pressure (MP) approach, namely efficient max pressure (Efficient-MP) using the EP to capture the traffic state; and (4) develop a general RL-based TSC algorithm template: efficient Xlight (Efficient-\textit{X}Light) under EP. 
Through comprehensive experiments on multiple real-world datasets in our traffic signal settings' protocol for TSC, we demonstrate that efficient pressure is complementary to traditional and RL-based modeling to design better TSC methods.
Our code is released on Github\footnote{\url{https://github.com/LiangZhang1996/Efficient\_XLight}}.

\end{abstract}

\section{Introduction}
\label{Sec:Intro}
Traffic signal control (TSC) plays a vital role in alleviating traffic congestion and improving road safety and transporation efficiency. A pre-determined cycle time plan for TSC was proposed back in 1958 \cite{webster1958}, which is still widely used in real traffic scenarios around the world. In 1963, a method estimated the mean and variance of the queue length in equilibrium for the phase time plan\cite{Miller1963}. Typically, traffic movement was modeled as a store and forward (SF) queuing system for TSC \cite{Mirchandani2001}.
In 2013, the max-pressure (MP) method\cite{Varaiya2013} was proposed  under the assumption of lanes' pressure (the difference between upstream and downstream queue length).
The MP method achieved satisfactory results among traditional approaches, as it maximizes the traffic network's throughput greedily.
These conventional approaches take the TSC as an optimization problem under several factors, such as vehicle arrival rate and lane capacity\cite{Varaiya2013}, which could model the traffic states\cite{Rios2017} for TSC.
However, traffic conditions are often affected by many other factors such as weather change and heterogeneous driver preferences. 

Reinforcement learning (RL)\cite{Sutton1998} methods have achieved outstanding results in computer games like Go and StarCraft II.
There is an emerging trend of employing RL for TSC\cite{Yau2017}. 
A typical RL-based approach is to model each intersection as an agent. The agent observes its traffic states (e.g., number of vehicles) and optimizes its reward (e.g., queue length) based on the feedback received from the traffic environment after it chooses an action (i.e., change the traffic signals).
The advantage of RL-based methods is that they can directly learn and adjust the strategies based on the feedback from the traffic environment.
Several RL-based studies, powered by deep learning (DL)\cite{LeCun2015}, have shown superior performance over traditional approaches\cite{Presslight2019,Yilmaz2020,HiLight2021}.

However, TSC is still a challenging issue considering that:
\begin{itemize} 
    \item Conventional approaches do not perform well due to a lack of capacity to adapt to dynamic traffic movements.
    \item RL-based methods are rarely deployed in the real TSC infrastructure due to the requirement of a long training process and significant computing resources.
    \item It is hard to precisely model a dynamic traffic movement.  
\end{itemize}
To address these challenges, we summarize our contributions in this paper as follows:
\begin{enumerate}
  \item We innovatively specify the traffic movement representation as a simple but efficient pressure of queue length in a traffic network, namely efficient pressure (EP); 
  \item A traffic signal settings protocol is built, including phase duration, signal phase number, and EP for TSC;
  \item A TSC approach is designed based on the traditional max pressure (MP) approach, namely efficient max pressure (Efficient-MP) using EP to capture the traffic state;
  \item A general RL-based TSC algorithm template is developed: efficient Xlight (Efficient-\textit{X}Light) under EP. 
\end{enumerate}

Through comprehensive experiments on multiple real-world datasets under our traffic signal settings protocol for TSC, we demonstrate that: (1) traditional MP outperforms most latest RL-based methods with similar performance as EP; (2) efficient-MP achieves a new state-of-the-art (SOTA) among all conventional and RL-based methods; and (3) the results of Efficient-\textit{X}Light are also significantly improved compared to the latest RL-based approaches.

\section{Related Works}
\label{Sec:Related}
\subsection{Conventional approaches}
In 1958, Webster et.al\cite{webster1958}  proposed the fixed-time control approach, which changes signal phases according to rules predefined for signal plans.
The fixed-time method mechanically controls the traffic signal without paying attention to vehicle queues, which could easily lead to unbalanced vehicle flow.
However, the fixed-time method is still widely used in real TSC for its simplicity.

Feedback signal control methods set the signals according to the traffic conditions on the road. 
SCOOT and SCATS\cite{SCATS1990}, consider the manually designed signal plans (set thresholds to determine if the signal should change) to optimize the alleviation of congestion.
In the store-and-forward (SF) queuing network model\cite{Mirchandani2001}, the traffic state is modeled as an SF vehicle queuing network. 
Feedback policies based on an SF model (queue measurements) are generated to control traffic signals adaptively.
The max pressure (MP) control\cite{Varaiya2013} was implemented at an intersection in a decentralized model.
The MP approach depends on the pressure, which represents the difference of queue length between upstream and downstream, to stabilize and maximize network throughput without requiring knowledge of the external arrivals (vehicle queue demands).
Thus, the MP method can automatically adapt to slow changes in traffic movement patterns, which have achieved good results among traditional methods.
\subsection{RL-related methods}
RL-related methods are designed for different application scenarios including single intersection control\cite{li2016traffic}, and multi-intersection control\cite{el2014design,rasheed2020deep}.
In 2019, FRAP\cite{FRAP2019} architecture for RL-based TSC, captured the competition relation between different traffic movements and achieves invariance to symmetry properties. 
Moreover, it leads to better solutions like (PressLight\cite{Presslight2019}, MPLight\cite{MPlight2020} and CoLight\cite{CoLight2019}, all based on FRAP) for TSC.
In 2021,  a hierarchical and cooperative RL method\cite{HiLight2021} to learn a high-level policy that optimized the objective locally by selecting among the sub-policies.
When all phases (different traffic movements) complying with the traffic rules are considered, the solution exploration space increases vastly.
The key to the RL-based approach is to reduce exploration space and explore different scenarios more effectively with minimal trials.
\subsection{Traffic movement representation}
Most TSC methods are feedback signal control under traffic states.
Traditional approaches describe traffic state with different models.
The SF method\cite{Mirchandani2001} represents the traffic as a vehicle queuing network, while the MP approach\cite{Varaiya2013} models the traffic movement as lanes' pressure.
The average queue length of all the lines is used for computing pressure\cite{Kouvelas2014}. However, there is not a precise formula to represent the traffic state.
Existing RL methods differ in terms of the traffic state representation of the traffic environment. 
Different kinds of states (features) from traffic environment have been adopted. They include the queue length\cite{Mannion2016}, average delay \cite{el2014design}, image features\cite{VanDer2016}, number of vehicles (NV)\cite{CoLight2019}, traffic movemnt  pressure (calculated with NV)\cite{MPlight2020}, traffic movement pressure (calculated with queue length) (pressure-queue)\cite{Varaiya2013} and travel time\cite{HiLight2021}.
Moreover, various traffic signal settings like 4-phase\cite{Presslight2019}, 8-phase\cite{MPlight2020} and 5-phase\cite{HiLight2021} are used for TSC methods. 
To achieve adaptive TSC, effective and efficient traffic state representation and traffic signal settings could be the key, rather than complex algorithms. Therefore, in this paper, efficient pressure (EP) is designed for computing pressure and representing the state of traffic, which is expected to be effective in TSC for multiple reasons.
 
Firstly, according to traffic rules, drivers should keep their lanes. However, vehicles often change lanes in the middle of an intersection before entering exiting lanes without breaking traffic rules. However, this behavior is usually neglected when presenting a traffic movement state. Secondly, the actual traffic movement is the queue length difference between entering and exiting lanes rather than lane to lane. The EP can completely depict the imbalance of entering lanes and exiting lanes. At last, the public datasets we choose have the lane-changing behaviors in the middle of an intersection, which supports our EP ideas to conduct experiments.

\section{Preliminary}
\label{Sec:Pre}
\subsection{Definition} 
\begin{figure}[h]
    \centering
    \includegraphics[width=1.02\linewidth]{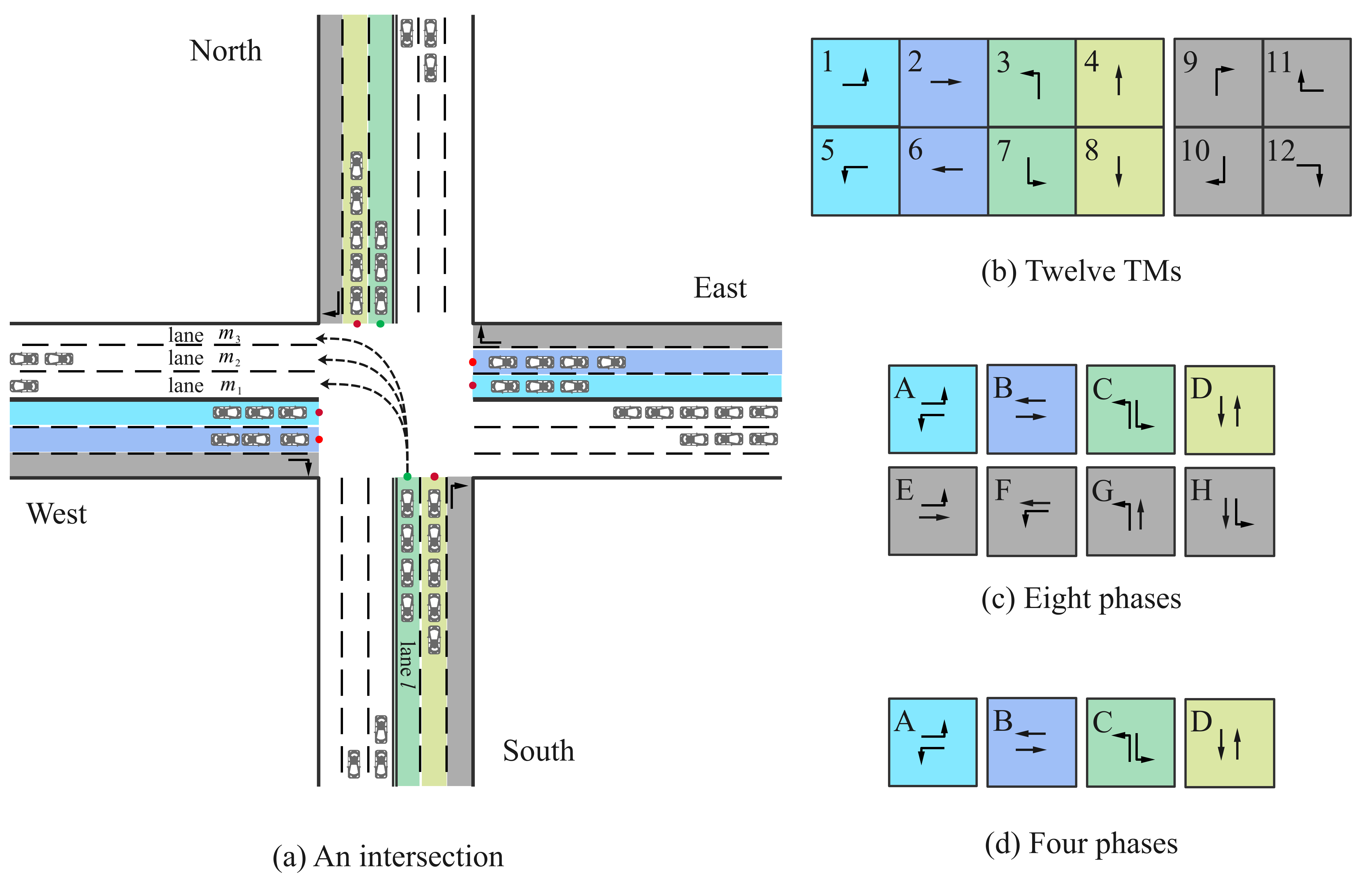}
    \caption{Illustration of an intersection and different phases.}
    \label{def}
        \vspace{-0.2cm}  
\end{figure}

To illustrate relevant definitions, we use a 4-approach crossroad as a typical intersection, as shown in Figure~\ref{def} (a). Each entering road has traffic movements of turn-left, go-straight, and turn-right. Anyhow, this example can be easily generalized and extended from 4-approach crossroads to different intersection structures.
 
\begin{itemize}
    \item Lane: 
    A lane $l$ is part of a road designated to be used by a single line of vehicles.

    \item  Intersection: An intersection  $I$ is a junction where several roads converge or cross. Each intersection has a set of entering lanes $\mathcal{L}^{in}$ and a set of exiting lanes $\mathcal{L}^{out}$.

    \item  Traffic networks: Each traffic network is a set of intersections $(I_1,...,I_N)$ connected by a set
     of roads ($(R_1,...,R_M)$ ), where $N$ is the number of intersections,  $M$ is the number of roads.
 
    \item  Traffic movement: A traffic movement (TM) represents vehicles traveling across an intersection from one entering lane $l$ to an exiting lane $m$, denoted as $(l,m)$. 
    
    As  Figure~\ref{def} (b) shows, a 4-approach intersection has twelve TMs, including four go-straight TMs, four "left-turn" TMs and four "right-turn" TMs (gray directions). According to the the traffic rules in most countries, vehicles can turn right regardless of the signal, but sometimes they need to stop at a red light. Therefore, there are usually eight adjustable TMs (\#1--\#8).
    \item Signal phase: Theoretically, one TM can be represented by one signal. But it is more efficient to use paired-TM (TM pairs) as a phase. A paired-TM phase  $s$ is a combination of two single signals (TMs), denoted as:
    \begin{equation} 
        s= \{(l,m),(k,v)\}  
    \end{equation} 
   
    wherein $(l,m)$ and $(k,v)$ are two TMs with entering lane $l,k$ and exiting lane $k,v$.  

    As Figure~\ref{def} (c) and Figure~\ref{def} (d) shows all single signals would combine eight or four effective signal pairs settings with non-conflicting signals (8-phase: 'A' to 'H' , 4-phase: 'A' to 'D'). 
        
    \item Traffic movement pressure: A TM pressure is the difference of queue length between the entering lane ($l$) and exiting lane ($m$) , denoted as: 
    \begin{equation} 
    p(l,m)= x(l)-x(m)
    \end{equation}
    wherein $x(l)$ and $x(m)$ represent the queue length at lanes $l$ and $m$, respectively.
        
    \item Phase pressure: The pressure of a  phase is the sum pressure of two TMs from one phase, denoted as $p(s)$:
    \begin{equation} 
    p(s)= x(l,m) + x(k,v)   
    \end{equation} 
    wherein $(l,m)$ and $(k,v)$ are two TMs of entering lanes $l,k$ and exiting lanes $k,v$, respectively. As Figure \ref{intersection} shows, the pressure of $p\{(l,m),(k,v)\}$ is calculated as: $x(l,m)+ x(k,v) = (4-1)+(3-5)=1$.

    \item Intersection pressure: The pressure of each intersection is the difference of queue length between all entering lanes ($\mathcal{L}_{i}^{in}$) and exiting lanes ($m \in \mathcal{M}_{i}^{out}$), denoted as:
    \begin{equation}  
    P_i = \sum x(l)-\sum x(m),(l \in \mathcal{L}_{i}^{in}, m \in \mathcal{M}_{i}^{out})
    \end{equation} 

    \item  Phase duration: The minimum duration for each phase signal is denoted as $t_{duration}$.
\end{itemize}
All notions are listed on Table~\ref{tab:list}.
\begin{table}[htb]
    \begin{tabular}{cl}
        \toprule
    Notation & Description\\
    \midrule
    $\mathcal{L}_{i}$  & set of lanes of intersection $i$\\
    $l, m, k, v$ & lane ($l,k$ for incoming; $m,v$ for outgoing)  \\ 
    $(l,m)$ &  a traffic movement from lane $l$ to $m$\\
    $(k,v)$ &  a traffic movement from lane $k$ to $v$\\
    $x(l)$ & queue length at lane $l$\\ 
    $p(l,m)$ &  pressure of a traffic movement $(l,m)$\\
    $s$ & phase which is a set of traffic movements\\
    $P_i$ & pressure of intersection $i$\\
    $p(s)$ & pressure of phase $s$\\
    $t_{duration}$ &  the minimum duration for each phase\\
   \midrule
    $\mathcal{\hat{L}}^{in}, \mathcal{\hat{M}}^{in}$  & lanes $\mathcal{\hat{L}}^{in}, \mathcal{\hat{K}}^{in}$ for incoming; \\
    
    $\mathcal{\hat{M}}^{out}, \mathcal{\hat{V}}^{out}$  & lanes $\mathcal{\hat{K}}^{in}, \mathcal{\hat{V}}^{in}$ for outgoing; \\
    
    $(\mathcal{\hat{L}}^{in},\mathcal{\hat{M}}^{out})$ &  An ETM  from lanes $\mathcal{\hat{L}}^{in}$ to $\mathcal{\hat{M}}^{out}$\\

    $\hat{p}_e(\mathcal{\hat{L}}^{in},\mathcal{\hat{M}}^{out})$ & EP from lanes $\mathcal{\hat{L}}^{in}$ to $\mathcal{\hat{M}}^{out}$ \\

    $\hat{p}_e(\mathcal{\hat{K}}^{in},\mathcal{\hat{V}}^{out})$ & EP from lanes $\mathcal{\hat{K}}^{in}$ to $\mathcal{\hat{V}}^{out}$ \\

    $s_e$  & an efficient signal phase \\

    $\hat{p}_e(s_e)$ & an EP for a phase $s_e$\\
    \bottomrule
    \end{tabular}
      \caption{List of notations in this paper.}
      \label{tab:list}
    \end{table}
    \vspace{-0.4cm}

\subsection{Problem description}
Multi-intersection TSC optimization: Traffic network consists of intersections $(I_1,...,I_N)$ and roads. Each intersection $I_i$ is controlled by an algorithm $A_i$ ($A_i$ could be a traditional or RL-based method). At time step $t$, $A_i$ makes an optimal decision to choose proper signal phase $s$ after analyzing its observation of the traffic state at the intersection. In this case, the time $T$ (average travel time) that vehicles spend in the traffic network can be minimized, denoted as:
\begin{equation} 
\min\sum_{i=1}^{N}T_i^{I_i(A_i)}
\end{equation} 
where $T_i^{I_i(A_i)}$ is the average time vehicles spending at intersection $I_i$ based on algorithm $A_i$, and $N$ is the number of intersections.

\section{Methods}
In this section, we introduce the newly proposed efficient pressure (EP), a traffic signal settings protocol, a TSC approach efficient max pressure (Efficient-MP), and a general RL-based TSC algorithm template: Efficient-\textit{X}Light.

\subsection{Efficient Pressure}
The pressure of an intersection is affirmative, while the pressure for traffic movement and phase is uncertain. There are several ways to compute the pressure (e.g., MPLight\cite{MPlight2020} using number of vehicles). Although traffic signal controls traffic movement rather than a lane, we find that calculating pressure from "lane to lane"  to "lanes to lanes" could lead to different results and TSC effects.

Inspired by the MP control\cite{Varaiya2013}, which realized minimal average travel time through balancing the queue length in the network, we design a simple but efficient traffic state representation, in other words, the "lanes to lanes" way to calculate pressure, namely efficient pressure (EP).

To depict the EP precisely, multiple new concepts based on basic definitions in the section~\ref{Sec:Pre} are presented as follows:
\begin{itemize}
    \item \textbf{Definition 1} - Effective traffic movement. 
    An effective traffic movement (ETM) happens when vehicles across an intersection from entering lanes to exiting lanes (as shown in Figure \ref{effective-movement}). The direction of vehicles traveling is in line with the TM defined.  

        \begin{figure}[h]
            \centering
            \includegraphics[width=0.8\linewidth]{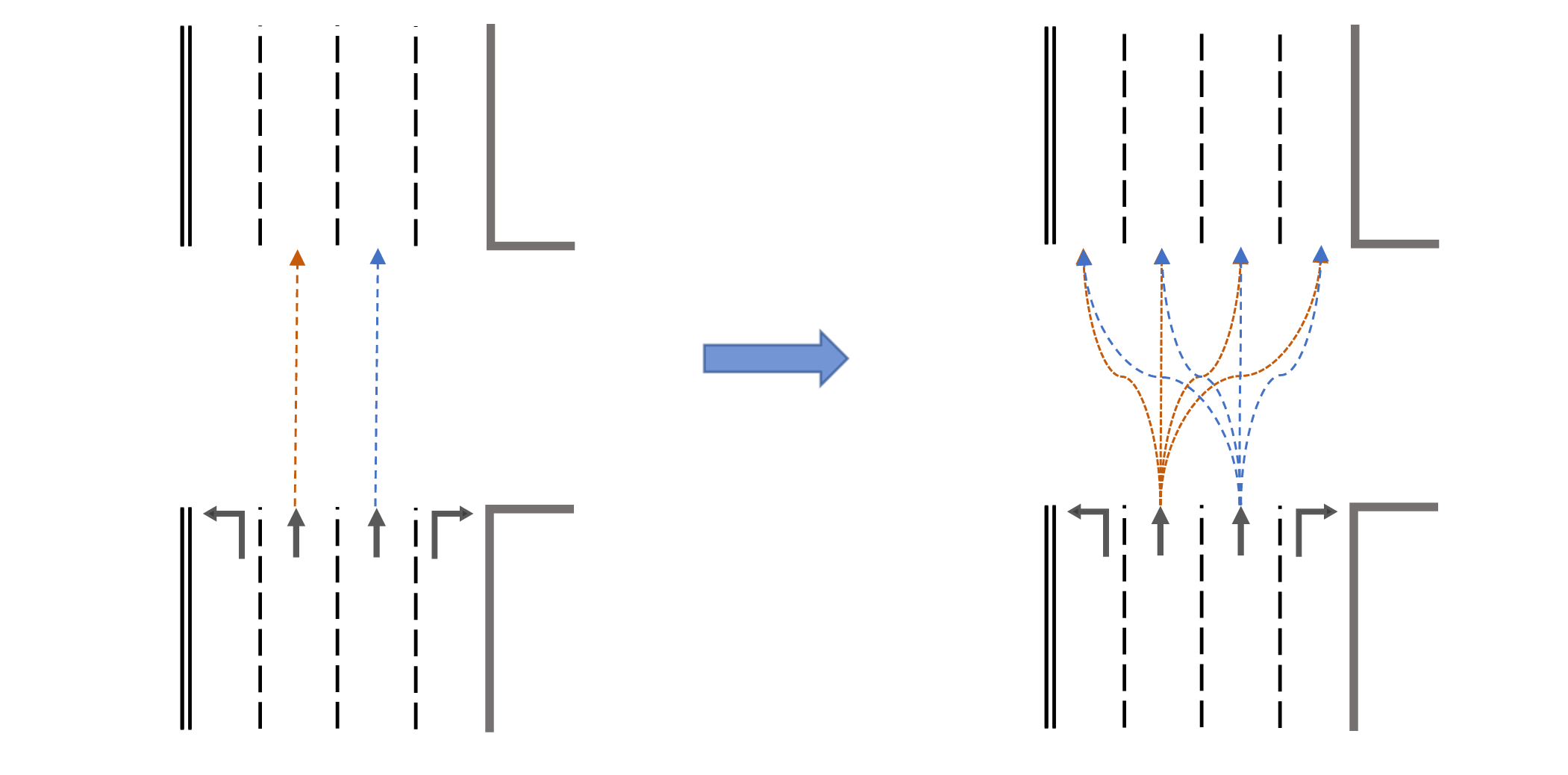}
            \caption{
                Illustration of "go-straight" ETM
            }
            \label{effective-movement}
              \vspace{-0.2cm}  
        \end{figure}

    \item \textbf{Definition 2} - Efficient pressure. The efficient pressure (EP) of one ETM is the difference between the average queue length on entering lanes and the average queue length on exiting lanes ($l_i\in \mathcal{\hat{L}}^{in},m_j\in \mathcal{\hat{M}}^{out}$).
    \begin{equation}
        \begin{aligned}
        \hat{p}_e(\mathcal{\hat{L}}^{in},\mathcal{\hat{M}}^{out}) & =   \frac{1}{L} \sum_{1}^{L}x({l_i})-  \frac{1}{M}\sum_{1}^{M}x(m_j)
        \end{aligned}
    \vspace{-0.2cm}  \end{equation}  
    where $\mathcal{\hat{L}}^{in}$ and $\mathcal{\hat{M}}^{out}$ ( $l_i\in \mathcal{\hat{L}}^{in},m_j\in \mathcal{\hat{M}}^{out}$) represent one ETM's upstream (entering lanes with number of $L$) and its downstream (exiting lanes with number of $M$), respectively.

    As Figure~\ref{def} (a) shows, the EP from $\mathcal{\hat{L}}^{in}$ (one lane $l $) to $\mathcal{\hat{M}}^{out}$
    $\mathcal{\hat{L}}^{in}$ (three lanes $m_1,m_2,m_3$). The efficient pressure in this case is $3=\frac{4}{1}-\frac{1+2+0}{3}$.
    \item \textbf{Definition 3} - Efficient signal phase. An efficient signal phase $s_e$ can be 8-phases or 4-phases, same as the signal phase definition, which combines two ETMs without conflicting signals. For simplicity, the direction and representation of one ETM are the same as one TM.
    \begin{equation} 
        s_e= \{(\mathcal{\hat{L}}^{in},\mathcal{\hat{M}}^{out}) ,(\mathcal{\hat{K}}^{in},\mathcal{\hat{V}}^{out})\} \\
    \end{equation} 
    where $(\mathcal{\hat{K}}^{in},\mathcal{\hat{V}}^{out})$ is the other ETM of entering lanes $\hat{K}^{in}$ and exiting lane $\mathcal{\hat{V}}^{out}$. 

    \item \textbf{Definition 4} - Phase efficient pressure. The EP of an efficient signal phase is the sum of the EP (two ETMs) from the phase, denoted as $\hat{p}_e(s)$:
    \begin{equation} 
    \hat{p}_e(s_e)= \hat{p}_e(\mathcal{\hat{L}}^{in},\mathcal{\hat{M}}^{out}) + \hat{p}_e(\mathcal{\hat{K}}^{in},\mathcal{\hat{V}}^{out})  
    \end{equation} 
\end{itemize}

\subsection{Basic protocol for TSC}
Now, PressLight\cite{Presslight2019}, FRAP\cite{FRAP2019}, and CoLight\cite{CoLight2019} have not yet depicted the specific protocol on signal duration or signal phase, which could significantly affect the final result.


We set phase duration as $t_{duration}=15$ under real-world datasets. $t_{duration}$ can be set pragmatically based on experiments upon other datasets. 
We aim to prove that, even with a fixed-time approach\cite{webster1958}, a proper signal setting could improve TSC performance significantly. Moreover, the MP method can even outperform current RL-related method MPLight\cite{MPlight2020}.

Although FRAP\cite{FRAP2019} stated that 8-phase was better than 4-phases regarding the results during rush hour, we choose 4-phase for the signal phase after comprehensive experiments regarding average travel time. Moreover, the 4-phase setting is simple. Based on MP\cite{Varaiya2013}, an efficient max pressure (Efficient-MP) algorithm under EP is developed.

Our traffic signal settings is not supposed to be universal for all datasets or conditions, though it is suited for JiNan and HangZhou  datasets (detail in experiments). Basic protocol for TSC should depict the phase duration and phase numbe.

\subsection{Efficient-MP}
The Efficient-MP method considers EP as a traffic state, while the other steps are the same as MP. At intersection $i$, for each phase $s_e\in \mathcal{S}_i$, $S_i$ is the set of signal phase. The pressure $\hat{p}_(s)$ is calculated by equation (8), then we select the phase with maximum pressure, denoted as: 
\begin{equation}
    \hat{s}_e = \arg \max (\hat{p_e}(s_e) |  s_e \in \mathcal{S}_i)
\end{equation}


\begin{algorithm}[H]
    \label{Efficient-MP}
	\caption{Efficient-MP}
	\textbf{Parameters}: Current phase time $t$, minimum phase duration $t_{duration}$

	\begin{algorithmic}
		\FOR{(time step)}
		\STATE $t = t+1$; 
		\IF{$t=t_{duration}$}
        \STATE Input: Get $\hat{p}(s_e)$ by equation (8) for each intersection;\\
        \STATE Output: Set the phase according to equation (9);
        
        \STATE $t=0$; 
		\ENDIF
		
		\ENDFOR
	\end{algorithmic}
\end{algorithm}
\vspace{-0.3cm}  
\subsection{Efficient-\textit{X}Light}
We develop an efficient RL-based methods template applying EP as the traffic state, namely Efficient-\textit{X}Light based on different RL models reported in literature recently.

The Efficient-\textit{X}Light template applies EP as traffic state and processes the original RL method, as shown in Algorithm 2. We also propose Efficient-PressLight, Efficient-CoLight, and  Efficient-MPLight using our Efficient-\textit{X}Light template based on the following: PressLight\cite{Presslight2019}, CoLight\cite{CoLight2019} and MPLight\cite{MPlight2020}, respectively. 
\begin{algorithm}[h]
   \label{Efficient-XLights}
	\caption{Efficient-XLight}
	\textbf{Parameter}: Current phase time $t$, minimum action duration $t_{duration}$
	\begin{algorithmic}
		\FOR{(time step)}
		\STATE $t=t+1$;
        \IF{$t=t_{duration}$} 
        \STATE Input: Get $\hat{p}(s_e)$ by equation (8) for each intersection;\\
        \STATE Output: Set the phase by \textit{X} RL model; 
        \STATE $t=0$;
		\ENDIF
		\ENDFOR
	\end{algorithmic}
\end{algorithm}
    \vspace{-0.2cm}  

Each agent observes the current phase and efficient pressure. At time $t$, each agent chooses a phase $\hat{s}$ as its action $a_t$, and the traffic signal will be set to phase $\hat{s}$. For Efficient-MPLight and Efficient-PressLight model, the reward is the intersection pressure $-|P_i|$; For Efficient CoLight, the reward is the intersection's queue length
$-\sum x(\bar{l}), \bar{l}\in \mathcal{L}_i^{in}$.

The Efficient-XLight is updated by the Bellman Equation:
\begin{equation}
	Q(s_t, a_t)  = R(s_r, a_t) + \gamma max Q(s_{t+1}, a_{t+1})
\end{equation}

Theoretically, we could build more RL models by Efficient-XLight. However, because the code PressLight\cite{Presslight2019}, CoLight\cite{CoLight2019} and MPLight\cite{MPlight2020} are all readily available, we implement these three first, without loss of validity of our conclusion.

\section{Experiments} 
We conduct experiments on an open-source simulator CityFlow\cite{CityFLow}, which has been widely used by multiple RL-based methods such as CoLight\cite{CoLight2019} and MPLight\cite{MPlight2020}.  

\subsection{Datasets} 

We use five real-world datasets in the experiments, three from JiNan and two from HangZhou in China. 
\begin{itemize}
\item \textbf{JiNan dataset:} The traffic network is $3\times4$. Each intersection is set to be a 4-approach intersection, with two 400-meters (East-West) long road segments and two 800-meters (South-North) long road segments. 
\item \textbf{HangZhou dataset:} The road network is $4\times4$. Each intersection is set to be a 4-approach intersection, with two 800-meter (East-West) long road segments and two 600-meter (South-North) long road segments. 
\end{itemize}

\subsection{Evaluation metrics} 
Travel time of each vehicle is the time discrepancy between entering and leaving a transportation area (in seconds). The average travel time of all cars in a traffic network is the most frequently used metric to evaluate the performance of TSC.

\begin{table*}[htb]
    \begin{center}
    \begin{tabular}{lccccc}
        \toprule
        \multirow{2}{*}{ Method } & \multicolumn{3}{c}{ JiNan } & \multicolumn{2}{c}{ HangZhou } \\
        \cline { 2 - 6 } & 1 & 2 & 3 & 1 & 2 \\
        \midrule
        FixedTime    & $428.11 (+56.27\%)$ & $368.77 (+50.29\%)$ & $383.01 (+55.82\%)$ & $495.57 (+71.75\%)$ & $406.65(+16.53\%)$ \\
        MP           & $273.96$ & $245.38$ & $245.81$ & $288.54$ & $348.98$ \\
        \midrule
        PressLight & $314.63 (+14.84\%) $ & $264.62 (+7.84\%)$ & $258.12 (+5.01\%) $ & $385.71 (+33.68\%) $ & $458.12 (+31.27\%)$ \\
        MPLight & $297.46 (+8.58\%)$ & $270.05 (+10.06\%)$ & $276.15 (+12.34\%)$ & $314.60 (+9.03\%) $ & $357.61 (+2.47\%) $ \\
        CoLight & {$\mathbf{272.06 (-0.69\%)}$} & ${252.44 (+2.88\%)}$ & ${249.56 (+1.53\%)}$ & ${297.02 (+2.94\%)}$ & {$\mathbf{347.27 (-0.49\%)}$} \\
        \midrule
        Efficient-MP & {$\mathbf{269.87 (-1.49\%)}$ }& {$\mathbf{239.75 (-2.29\%)}$} & {$\mathbf{240.03 (-2.35\%)}$ }& {$\mathbf{284.44 (-1.42\%)}$} & {$\mathbf{327.62 (-6.12\%)}$} \\
		Efficient-PressLight & $278.96 (+1.83\%)$ & $253.10 (+3.15\%)$ & $253.16 (+2.99\%)$ & $312.32 (+8.24\%)$ & $414.32 (+18.72\%)$ \\
		Efficient-MPLight &  {$\mathbf{261.81 (-4.43\%)}$} &  {$\mathbf {241.35 (-1.64\%)}$ }&  {$\mathbf { 238.80 (-2.85\%)}$ }&  {$\mathbf {284.49 (-1.40\%)}$} &  {$\mathbf{3 2 1 . 0 8 (-7.99\%)}$ }\\
        Efficient-CoLight & {$\mathbf{2 5 6 . 8 4 (-6.25\%)}$} &  {$\mathbf{2 3 9 . 5 8 (-2.36\%)}$ }&  {$\mathbf{2 3 6 . 7 2(-3.70\%)}$} &  {$\mathbf{2 8 2 . 0 7 (-2.24\%)}$} &  {$\mathbf{324.27 (-7.08\%)}$} \\
        \bottomrule
    \end{tabular}
       \caption{Performance comparison (travel time in seconds) of different methods evaluated on JiNan and HangZhou datasets.}
       \label{tab:results}
    \end{center}
    \vspace{-0.4cm}  
    \end{table*}
\subsection{Methods for Comparison}
To evaluate our models, we compare our method with the following baseline approaches. 

\textbf{Transportation Methods}:
\begin{itemize}
\item \textbf{Fixed-Time}\cite{webster1958}: a policy gives a fixed cycle length with predefined green time among all phases.

\item \textbf{MP}\cite{Varaiya2013}: the MP control selects the phase that has the maximum pressure. 
\end{itemize}

\textbf{RL Methods}
\begin{itemize}
\item \textbf{PressLight}\cite{Presslight2019}: the PressLight uses the current phase, number of vehicles in entering lanes, and number of vehicles in exiting lanes as traffic state, and intersection pressure as a reward.
\item \textbf{MPLight}\cite{MPlight2020}: it uses current phase and traffic movement pressure (number of vehicles) as traffic state, and intersection pressure as a reward.

\item \textbf{CoLight}\cite{CoLight2019}: the CoLight uses the current phase and number of vehicles as traffic state and intersection queue length as a reward. 
\end{itemize}

\textbf{Our Proposed Methods}

\begin{itemize}
\item\textbf{Efficient-MP}: the Efficient-MP selects the phase with the maximum efficient pressure. 
\item \textbf{Efficient-PressLight}: PressLight based model, uses the current phase, EP as a traffic state, and intersection pressure as a reward.
 
\item \textbf{Efficient-MPLight}: an MPLight based model, uses current phase and EP as traffic state, and intersection pressure as a reward.

\item \textbf{Efficient-CoLight}: CoLight based model, uses current phase and EP as traffic state, and intersection queue length as a reward. 
\end{itemize}

\subsection{Experiment settings}
All the RL-related methods are trained with the same hyper-parameters: learning-rate, replay buffer size, sample size. Each episode is a 60-minute simulation, and we report one result set as the average of the last 10 episodes of testing. The final result set is the average of three independent results.
According to our traffic signal protocol, we pragmatically set phase duration as $t=15$ seconds and signal phase as 4-phases after observing several experiments for all methods. 
Each green signal is followed by a three-second yellow signal and two-second all red time.

\subsection{Results}
\subsubsection{Phase duration}
\begin{figure}[h]
    \includegraphics[width=1.05\linewidth]{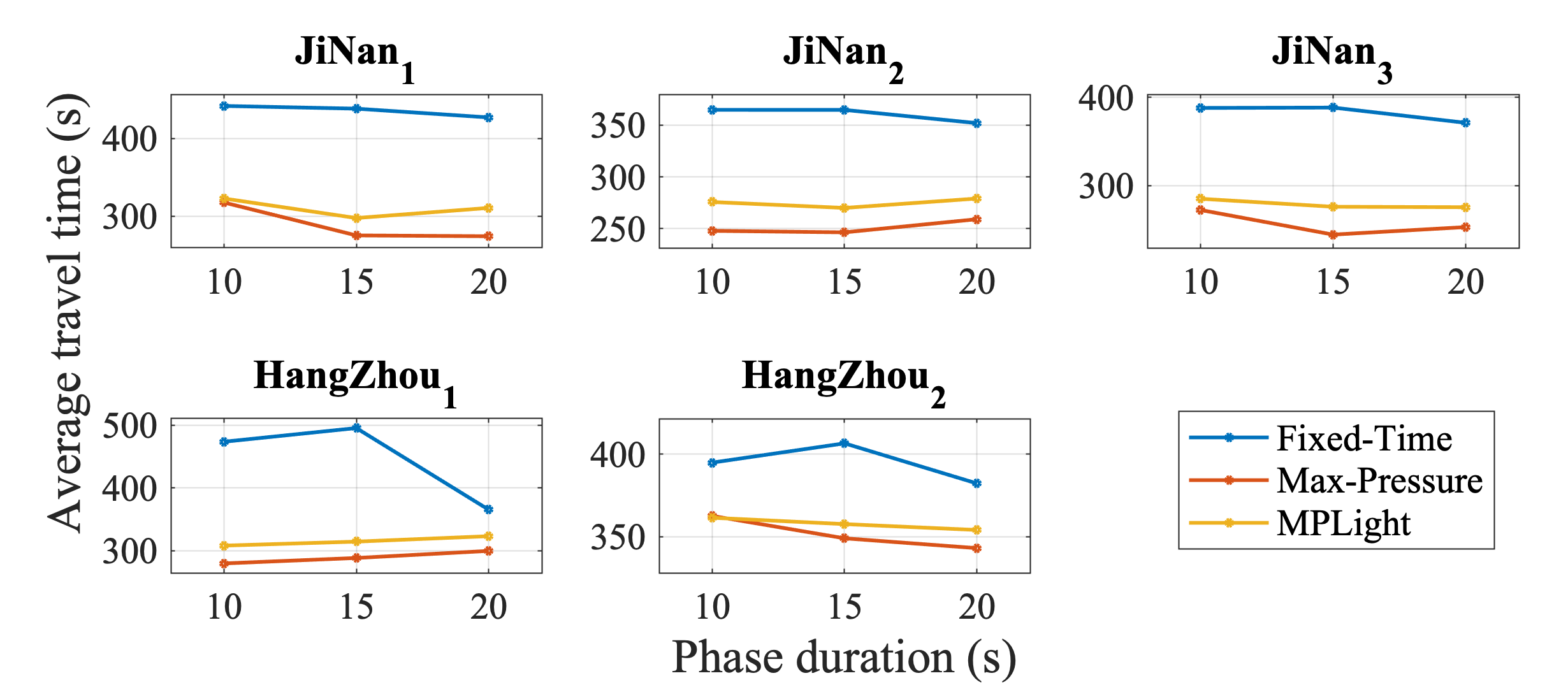}
       \caption{Performance (average travel time in seconds) comparison of different methods with $t_{duration}=$ 10, 15 and 20, respectively.}
    \label{fig:duration}
        \vspace{-0.2cm}  
\end{figure}

As Figure \ref{fig:duration} shows, TSC methods are affected by the phase duration.
A unified phase duration is necessary for model comparison, and we set $t_{duration}=15$ as our and baseline methods.
Moreover, we find that the traditional MP model can outperform the lasted RL approach like MPLight for a proper phase duration. 

\subsubsection{Signal Phase}
We conducted experiments on two signal phase scenarios for different methods evaluated on JiNan and HangZhou real-world datasets. As Table~\ref{tab:phase} shows, 4-phase would be better. In addition, 4-phase is more common in the real-world. Thus, we use 4-phase as our traffic signal settings protocol.
\begin{table}[h]
        \begin{center}
        \begin{tabular}{lccc}
        \toprule
         { Method } &{ Phase } & {JiNan } & {HangZhou} \\
        \midrule
        \multirow{2}*{MP}  & 4-phase & $\mathbf{255.05} $ & ${318.76} $ \\

         &8-phase & $261.07$ & $\mathbf{306.78}$ \\
        \midrule
        \multirow{2}*{MPLight}  & 4-phase & $\mathbf{281.22}$ & $\mathbf{336.11} $ \\
        &8-phase & $302.00 $ & $337.71 $ \\
       \midrule
       \multirow{2}*{CoLight}  & 4-phase & $\mathbf{258.02} $ & ${322.15} $ \\
       &8-phase & $271.3 $ & $\mathbf{315.17}$ \\
        \bottomrule
    \end{tabular}
    \caption{Performance (average travel time in seconds) comparison of different methods with different signal phases.}
    \label{tab:phase}
    \end{center}
       \vspace{-0.4cm}     
\end{table}

\subsubsection{Model Performance}

Based on our traffic signal settings protocol, the  phase duration  is set as 15s and signal is set as 4-phase. Experiments are conducted on JiNan and HangZhou datasets.
As Table~\ref{tab:results} shows: (1) the traditional MP approach outperforms the current methods such as PressLight, MPLight, and Colight; (2) Our Efficient-MP achieves a new state-of-the-art (SOTA); (3) the results of our RL-related methods generated by Efficient-\textit{X}Light template are also largely improved if compared to their original models.

To illustrate the importance of the EP in one TSC method further, we choose CoLight as the base algorithm and use four different representations of the traffic state. State $1$: number of vehicles (CoLight used); State $2$: TM pressure calculated by number of vehicles (MPLight used); State $3$: TM pressure calculated by queue length (MP used); State $4$: The EP. As Table \ref{tab:EP} shows, we demonstrate that EP is the most efficient way to represent the traffic state compared to others. 

We also demonstrate that more complex approaches can not guarantee better results. Our Efficient-MP has only the $\mathcal{O}(1)$ complexity, even not necessary to be compared to long-training RL-related approaches. For now, the EP is likely the key to traffic movement representation. 
\begin{table}[h]
   
        \begin{center}
        \begin{tabular}{cccccc}
        \toprule 
         \multirow{2}{*}{ State} & \multicolumn{3}{c}{ JiNan } &\multicolumn{2}{c}{HangZhou}\\
        \cline { 2 - 6 }  & 1 & 2 & 3 &1 &2 \\
        \midrule
         1 & $272.06 $ & $252.44 $ & $249.56 $ & $297.02$ &$347.27$\\
        2  & $274.47 $ & $253.71 $ & $251.43 $ & $298.76$& $347.71$ \\
        3 & $260.71 $ & $240.11 $ & $237.65 $ & $282.38$&$\mathbf{319.20}$\\
        4  & $\mathbf{256.84} $ & $\mathbf{239.58} $ & $\mathbf{236.72} $ & $\mathbf{282.07}$ & $324.27$\\
        \bottomrule
    \end{tabular}
      \caption{Performance (average travel time  in seconds) comparison of different representations of traffic state in the method of CoLight.}
       \label{tab:EP}
    \end{center}
      \vspace{-0.6cm}  
\end{table}

\section{Conclusion}
In this paper, we try to find the key to traffic movement representation and demonstrate that the traffic state is one of the important factors for constructing a less training, lower complexity, and adaptive for TSC. We design efficient pressure (EP) as the traffic movement representation as an efficient and straightforward pressure of vehicle queues in a traffic network. We build a traffic signal settings protocol, including phase duration, signal  phase number. In addition, We propose the traditional method based on Efficient-MP and RL-based template Efficient-\textit{X}Light under the EP model. The experimental results demonstrate our approaches are efficient.

In further research, we will continue to optimize the EP by considering more information from the traffic movement. Simultaneously, Efficient-CoLight performs the best, though the MP performs almost the same level. We will propose a more effective RL-based algorithm for adaptive TSC.

\bibliographystyle{named}
\bibliography{ijcai22}

\begin{thebibliography}{}

\bibitem[\protect\citeauthoryear{Chen \bgroup \em et al.\egroup
  }{2020}]{MPlight2020}
Chacha Chen, Hua Wei, Nan Xu, Guanjie Zheng, Ming Yang, Yuanhao Xiong, Kai Xu,
  and Zhenhui Li.
\newblock Toward a thousand lights: Decentralized deep reinforcement learning
  for large-scale traffic signal control.
\newblock {\em Proceedings of the AAAI Conference on Artificial Intelligence},
  34(04):3414--3421, 2020.

\bibitem[\protect\citeauthoryear{El-Tantawy \bgroup \em et al.\egroup
  }{2014}]{el2014design}
Samah El-Tantawy, Baher Abdulhai, and Hossam Abdelgawad.
\newblock Design of reinforcement learning parameters for seamless application
  of adaptive traffic signal control.
\newblock {\em Journal of Intelligent Transportation Systems}, 18(3):227--245,
  2014.

\bibitem[\protect\citeauthoryear{Kouvelas \bgroup \em et al.\egroup
  }{2014}]{Kouvelas2014}
Anastasios Kouvelas, Jennie Lioris, S.~Alireza Fayazi, and Pravin Varaiya.
\newblock Maximum pressure controller for stabilizing queues in signalized
  arterial networks.
\newblock {\em Transportation Research Record: Journal of the Transportation
  Research Board}, 2421:133--141, 2014.

\bibitem[\protect\citeauthoryear{LeCun~Y. and G}{2015}]{LeCun2015}
Bengio~Y. LeCun~Y. and Hinton G.
\newblock Deep learning.
\newblock {\em Nature}, 521:436–444, 2015.

\bibitem[\protect\citeauthoryear{Li~Li and Feiyue}{2016}]{li2016traffic}
Lv~Yisheng Li~Li and Wang Feiyue.
\newblock Traffic signal timing via deep reinforcement learning.
\newblock {\em IEEE/CAA Journal of Automatica Sinica}, 3(3):247--254, 2016.

\bibitem[\protect\citeauthoryear{Lowrie}{1990}]{SCATS1990}
P.~R. Lowrie.
\newblock Scats, sydney co-ordinated adaptive traffic system: A traffic
  responsive method of controlling urban traffic.
\newblock {\em Roads and Traffic Authority NSW}, 1990.

\bibitem[\protect\citeauthoryear{Mannion \bgroup \em et al.\egroup
  }{2016}]{Mannion2016}
Patrick Mannion, Jim Duggan, and Enda Howley.
\newblock {\em An Experimental Review of Reinforcement Learning Algorithms for
  Adaptive Traffic Signal Control}, pages 47--66.
\newblock 05 2016.

\bibitem[\protect\citeauthoryear{Miller}{1963}]{Miller1963}
Alan~J. Miller.
\newblock Settings for fixed-cycle traffic signals.
\newblock {\em Journal of the Operational Research Society}, 14(4):373–386,
  1963.

\bibitem[\protect\citeauthoryear{Mirchandani and Head}{2001}]{Mirchandani2001}
Pitu Mirchandani and Larry Head.
\newblock A real-time traffic signal control system: Architecture, algorithms,
  and analysis.
\newblock {\em Transportation Research Part C: Emerging Technologies},
  9:415--432, 12 2001.

\bibitem[\protect\citeauthoryear{Rasheed \bgroup \em et al.\egroup
  }{2020}]{rasheed2020deep}
Faizan Rasheed, Kok-Lim~Alvin Yau, and Yeh-Ching Low.
\newblock Deep reinforcement learning for traffic signal control under
  disturbances: A case study on sunway city, malaysia.
\newblock {\em Future Generation Computer Systems}, 109:431--445, 2020.

\bibitem[\protect\citeauthoryear{Rios-Torres and Malikopoulos}{2017}]{Rios2017}
Jackeline Rios-Torres and Andreas~A. Malikopoulos.
\newblock A survey on the coordination of connected and automated vehicles at
  intersections and merging at highway on-ramps.
\newblock {\em IEEE Transactions on Intelligent Transportation Systems},
  18(5):1066--1077, 2017.

\bibitem[\protect\citeauthoryear{Sutton and Barto}{2018}]{Sutton1998}
Richard~S. Sutton and Andrew~G. Barto.
\newblock {\em Reinforcement Learning: An Introduction}.
\newblock The MIT Press, second edition, 2018.

\bibitem[\protect\citeauthoryear{Van~der Pol and Oliehoek}{2016}]{VanDer2016}
Elise Van~der Pol and Frans~A. Oliehoek.
\newblock Coordinated deep reinforcement learners for traffic light control.
\newblock In {\em NIPS'16 Workshop on Learning, Inference and Control of
  Multi-Agent Systems}, 2016.

\bibitem[\protect\citeauthoryear{Varaiya}{2013}]{Varaiya2013}
Pravin Varaiya.
\newblock The max-pressure controller for arbitrary networks of signalized
  intersections.
\newblock {\em Advances in Dynamic Network Modeling in Complex Transportation
  Systems}, pages 27--66, 2013.

\bibitem[\protect\citeauthoryear{Webster}{1958}]{webster1958}
F.~V. Webster.
\newblock Traffic signal settings.
\newblock Technical Report~39, 1958.

\bibitem[\protect\citeauthoryear{Wei \bgroup \em et al.\egroup
  }{2019a}]{Presslight2019}
Hua Wei, Chacha Chen, Guanjie Zheng, Kan Wu, Vikash Gayah, Kai Xu, and Zhenhui
  Li.
\newblock Presslight: Learning max pressure control to coordinate traffic
  signals in arterial network.
\newblock In {\em Proceedings of the 25th ACM SIGKDD International Conference
  on Knowledge Discovery \& Data Mining}, pages 1290--1298, 2019.

\bibitem[\protect\citeauthoryear{Wei \bgroup \em et al.\egroup
  }{2019b}]{CoLight2019}
Hua Wei, Nan Xu, Huichu Zhang, Guanjie Zheng, Xinshi Zang, Chacha Chen, Weinan
  Zhang, Yanmin Zhu, Kai Xu, and Zhenhui~Jessie Li.
\newblock Colight: Learning network-level cooperation for traffic signal
  control.
\newblock {\em Proceedings of the 28th ACM International Conference on
  Information and Knowledge Management}, 2019.

\bibitem[\protect\citeauthoryear{Xu \bgroup \em et al.\egroup
  }{2021}]{HiLight2021}
Bingyu Xu, Yaowei Wang, Zhaozhi Wang, Huizhu Jia, and Zongqing Lu.
\newblock Hierarchically and cooperatively learning traffic signal control.
\newblock In {\em Proceedings of the AAAI Conference on Artificial
  Intelligence}, volume~35, pages 669--677, 2021.

\bibitem[\protect\citeauthoryear{Yau \bgroup \em et al.\egroup
  }{2017}]{Yau2017}
{Kok Lim Alvin} Yau, Junaid Qadir, {Hooi Ling} Khoo, {Mee Hong} Ling, and Peter
  Komisarczuk.
\newblock A survey on reinforcement learning models and algorithms for traffic
  signal control.
\newblock {\em ACM Computing Surveys}, 50(3), 2017.

\bibitem[\protect\citeauthoryear{Yilmaz}{2020}]{Yilmaz2020}
Yasin Yilmaz.
\newblock Deep reinforcement learning for intelligent transportation systems: A
  survey.
\newblock {\em IEEE Transactions on Intelligent Transportation Systems}, pages
  1--22, 2020.

\bibitem[\protect\citeauthoryear{Zhang \bgroup \em et al.\egroup
  }{2019}]{CityFLow}
Huichu Zhang, Yaoyao Ding, and Weinan Zhang.
\newblock Cityflow: A multi-agent reinforcement learning environment for large
  scale city traffic scenario.
\newblock {\em In the Web Conference 2019 - Proceedings of the World Wide Web
  Conference (WWW)}, page 3620–3624, 2019.

\bibitem[\protect\citeauthoryear{Zheng \bgroup \em et al.\egroup
  }{2019}]{FRAP2019}
Guanjie Zheng, Yuanhao Xiong, Xinshi Zang, Jie Feng, Hua Wei, Huichu Zhang,
  Yong Li, Kai Xu, and Zhenhui Li.
\newblock Learning phase competition for traffic signal control.
\newblock pages 1963--1972, 11 2019.

\end{thebibliography}

\end{document}